\documentclass[10pt,twocolumn,letterpaper]{article}

\usepackage{iccv}
\usepackage{times}
\usepackage{epsfig}
\usepackage{graphicx}
\usepackage{amsmath}
\usepackage{amssymb}
\usepackage[ruled]{algorithm2e}
\usepackage[table, xcdraw, dvipsnames, svgnames, x11names]{xcolor}
\usepackage{amssymb}
\usepackage{colortbl}
\usepackage{enumitem}
\usepackage[misc]{ifsym}

\usepackage[breaklinks=true,bookmarks=false,colorlinks]{hyperref}

\iccvfinalcopy 

\def\iccvPaperID{3898} 

\ificcvfinal\fi

\begin{document}

\title{Excavating the Potential Capacity of Self-Supervised Monocular Depth Estimation}

\author{Rui Peng \qquad Ronggang Wang \textsuperscript{\Letter} \qquad Yawen Lai \qquad Luyang Tang \qquad Yangang Cai\\
School of Electronic and Computer Engineering, Peking University\\
{\tt\small \{ruipeng,tly926\}@stu.pku.edu.cn \{alanlyawen,caiyangang\}@pku.edu.cn rgwang@pkusz.edu.cn}
}

\maketitle
\ificcvfinal\thispagestyle{empty}\fi

\begin{abstract}
   Self-supervised methods play an increasingly important role in monocular depth 
   estimation due to their great potential and low annotation cost. To close the gap with 
   supervised methods, recent works take advantage of extra 
   constraints, \eg, semantic segmentation. However, these methods will inevitably 
   increase the burden on the model. In this paper, we show theoretical and empirical 
   evidence that the potential capacity of self-supervised monocular depth estimation can be 
   excavated without increasing this cost. In particular, we propose {\bf(1)} a novel data 
   augmentation approach called data grafting, which forces the model to explore more cues to 
   infer depth besides the vertical image position, {\bf(2)} an exploratory self-distillation 
   loss, which is supervised by the self-distillation label generated by our new 
   post-processing method - selective post-processing, and {\bf(3)} the full-scale network, 
   designed to endow the encoder with the specialization of depth estimation task and 
   enhance the representational power of the model. Extensive experiments show that our 
   contributions can bring significant performance improvement to the baseline with even less 
   computational overhead, and our model, named {\bf \em EPCDepth}, surpasses the previous 
   state-of-the-art methods even those supervised by additional constraints. 
   Code is available at \url{https://github.com/prstrive/EPCDepth}.
   
\end{abstract}

\section{Introduction} \label{sec:intro}

\begin{figure}[t]
   \begin{center}
      \includegraphics[trim={15cm 10cm 17cm 4.5cm},clip,width=1.0\linewidth]{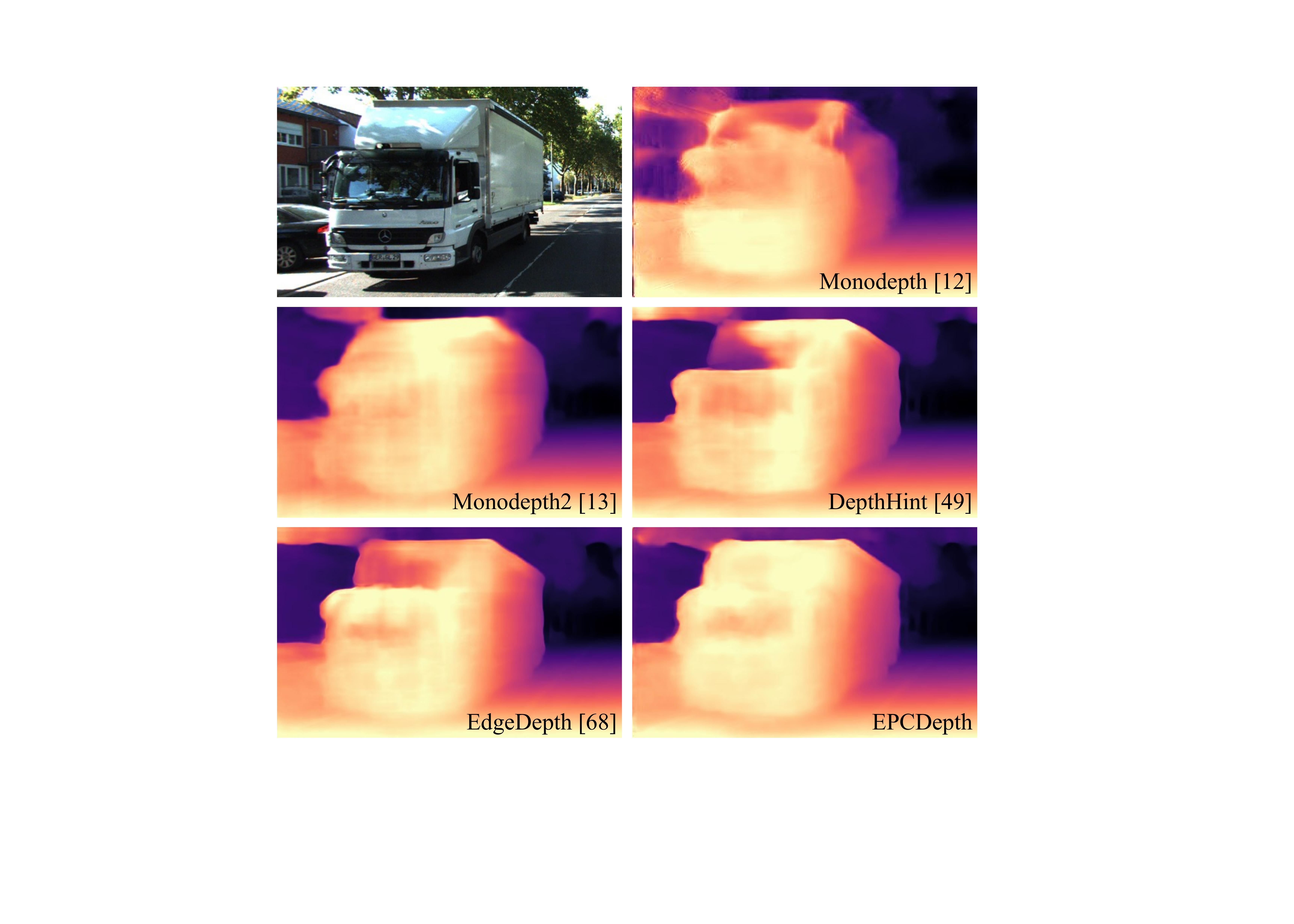}
   \end{center}
      \caption{{\bf Depth estimation from a single image.} Our model ({\bf EPCDepth}), 
      trained only on stereo data, performs the best and produces the sharpest and most 
      complete result with the least computational cost.}
   \label{fig:head}
\end{figure}

Depth estimation has always been a fundamental problem of computer vision, which 
dominates the performance of various applications, \eg, virtual reality, autonomous 
driving, robotics, \etc. As the cheapest solution, monocular depth estimation (MDE) 
has made considerable progress due to the evolvement of Convolution Neural 
Networks \cite{Krizhevsky2012,Simonyan2015,Szegedy2015,He2016}.  
However, most existing state-of-the-art approaches rely on supervised training 
\cite{Eigen2014,Eigen2015,Fu2018,Lasinger2020,bhat2021adabins}, 
whose training datasets collection is a cumbersome and formidable challenge. 
As an alternative, self-supervised methods eliminate the need for ground-truth 
depth through recasting depth estimation as the reconstruction problem among stereo 
images \cite{Garg2016,Godard2017,Watson2019,Zhu2020}, monocular 
video \cite{Zhou2017,Chen2019,Shu2020,Lyu2021} or a combination of 
both \cite{Zhan2018,Godard2019}.

In terms of performance alone, recent works have shown that the gap between 
self-supervision and full-supervision has made a de facto reduction. 
But on the other hand, this reduction largely benefits from the sophisticated model 
architecture and extra constraints from external modalities, \eg, 
semantic segmentation \cite{Chen2019,Klingner2020,Zhu2020,Guizilini2020Sem}, 
optical flow \cite{Yin2018,Ranjan2019}, depth normal \cite{Yang2018}, \etc. 
Apparently, these factors substantially increase the burden of the model 
training and run counter to the concept of self-supervision to some extent. 
In this paper, we show the potential of self-supervised monocular depth 
estimation even without these additional constraints from three aspects: 
{\bf data augmentation}, {\bf self-distillation}, and {\bf model architecture}.

Generally, the closer the projection on the image is to the lower boundary, the 
smaller the depth of the object. This feature of vertical image position has been proven 
to be the main cue adopted by the MDE model to infer depth, while the apparent size 
and other cues that humans will rely on are ignored \cite{VanDijk2019}. 
We conjecture that the reason is that in the traditional training mechanism that 
takes the entire image as input, the feature of vertical image position exists in 
almost every training sample, while the number of samples for other cues is 
relatively small, which leads to a long-tailed distribution on cues. Obviously, 
this kind of paranoia tends to damage the generalization ability of the model. 
To solve this, we propose a novel data augmentation method called 
{\bf \em Data Grafting}, which breaks this dilemma by vertically grafting a certain 
proportion from another image to appropriately weaken the relationship between 
depth and vertical image position. 
Moreover, there is another fact that the precision of different scales 
output by the multi-scale network is inconsistent at different pixels, and this motivates 
us to generate better disparity maps as pseudo-labels to realize the self-distillation 
of the model. Concretely, we propose {\bf \em Selective 
Post-Processing} (SPP) to select the best prediction for each pixel among all scales 
according to the reconstruction error, which is inspired by the availability of all 
views during training, and the similar idea has been proven effective in the field of 
multi-view stereo \cite{yi2020pyramid}. Finally, we extend the traditional 
multi-scale network to the full-scale network by inserting prediction modules not only 
on the decoder but also on the encoder to advance the specialization of depth 
prediction from decoder to encoder and absorb the representational power of the model. 
The superior result of our model is shown in Figure 
\ref{fig:head}.

To summarize, our main contributions are listed below in fourfold:

\begin{itemize}[leftmargin=*]
\item We introduce a conceptually simple but empirically efficient data 
augmentation approach, which enables the model to learn more effective cues 
besides the vertical image position.
\item We apply self-distillation to MDE for the first time without any auxiliary network 
and generate better pseudo-labels based on our training-oriented selective 
post-processing method.
\item We propose a more efficient full-scale network to strengthen the constraints 
on the model and enhance the encoder's specificity of depth estimation.
\item Without bells and whistles, we achieve state-of-the-art 
performance within self-supervised methods even compared to those 
high-performance models that are trained by extra constraints.
\end{itemize}

\begin{figure*}
   \begin{center}
      \includegraphics[trim={3cm 7cm 3cm 12cm},clip,width=0.95\linewidth]{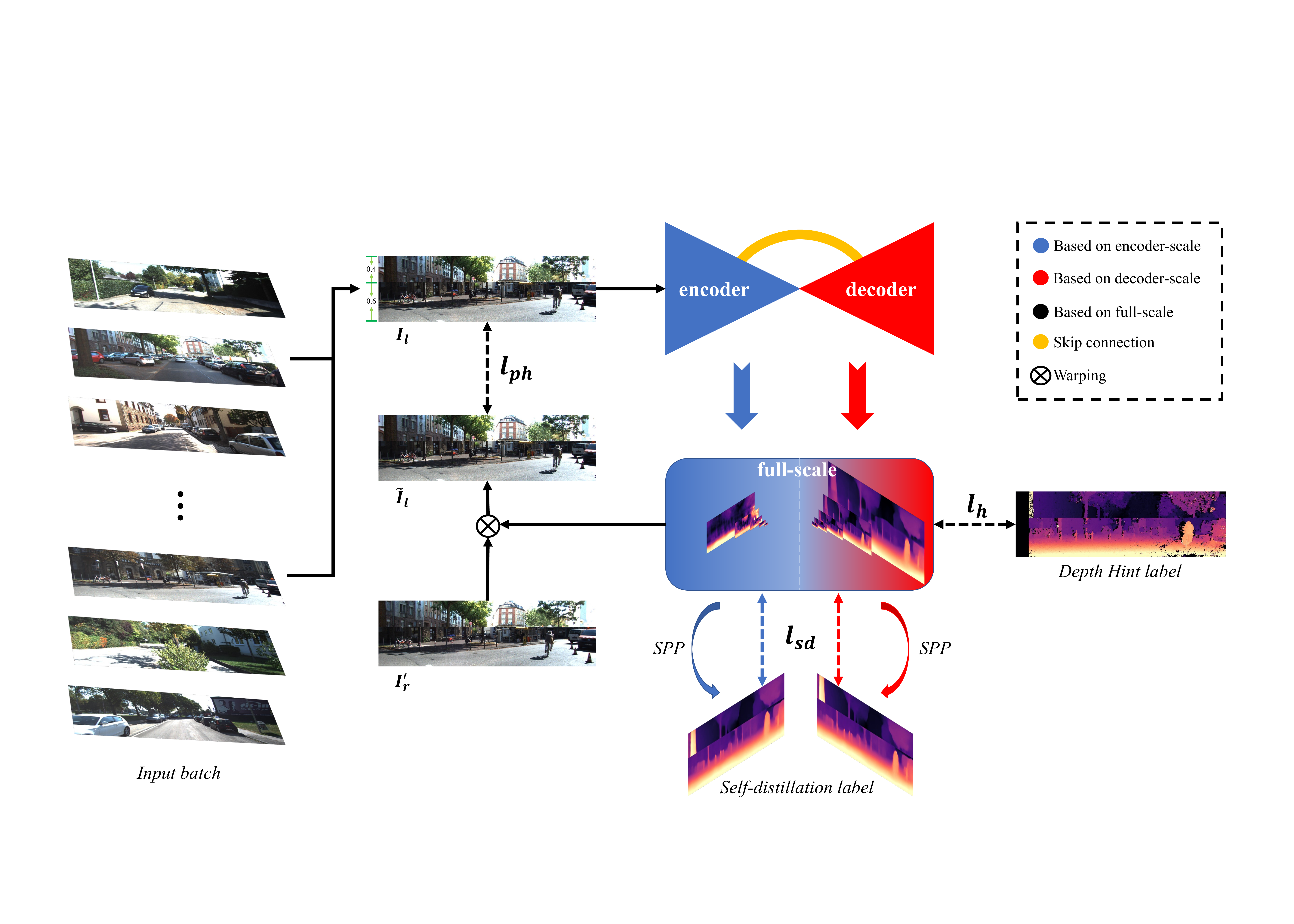}
   \end{center}
   \caption{{\bf Framework illustration.} The proposed approach is mainly composed of three 
   procedures. The input batch data is first refactored by data grafting, and here we take 
   the grafting ratio of 0.6 as an example. Immediately after that, the full-scale network 
   will estimate the disparity map at all scales, which means that not only the decoder 
   but also the encoder will infer the disparity. Finally, the full-scale disparity will 
   be used to generate the self-distillation label through selective post-processing for 
   the encoder and decoder scale separately and calculate the loss $l_{sd}$. Meanwhile, 
   the model will be trained with the assistance of photometric loss $l_{ph}$ and depth 
   hint loss $l_{h}$, and it is worth noting that these losses are executed on all scales.}
   \label{fig:framework}
\end{figure*}

\section{Related Works}

\noindent
{\bf Self-Supervised Monocular Depth Estimation.} The depth is predicted as 
an intermediate in self-supervised MDE to synthesize the reconstructed view from 
the source view, and the photometric loss between the target view 
and the reconstructed view is calculated as the target of minimization. There are 
mainly two kinds of self-supervised methods: trained by synchronized stereo 
images \cite{Garg2016,Godard2017,Pilzer2019,Watson2019,Zhu2020} or 
monocular video \cite{Zhou2017,Yin2018,Chen2019,Shu2020}. For the first category, 
the model with known relative placement only needs to 
predict the disparity, that is, the inverse of the depth. For the second category, 
additional predictions of the relative pose of the camera are required. Recently, 
abundant works have improved the performance of self-supervised MDE 
through new loss function 
\cite{Garg2016,Watson2019,Godard2019,Shu2020,Zhu2020}, new architecture 
\cite{Pilzer2019,Zhou2019,Yu2020,Guizilini2020,Lyu2021} and new supervision from 
extra constraints 
\cite{Yang2018,Yin2018,Ranjan2019,Chen2019,Klingner2020,Zhu2020, Guizilini2020Sem}.

In this paper, we further excavate the potential capacity of self-supervised 
MDE with the realization of training on stereo images.

\noindent
{\bf Self-Distillation.} Knowledge distillation is a pioneering work 
to transfer knowledge from powerful teacher networks to student 
networks using the softmax output \cite{Hinton2015}, intermediate feature 
\cite{Romero2015,Heo2019}, attention \cite{Zagoruyko2017,Hou2019}, 
relationship \cite{Yim2017,Park2019,Peng2019}, \etc. 
Self-distillation is a special case where the model 
itself is used as a teacher. Intuitively, the model can be 
distilled by the same model trained previously \cite{Furlanello2018}, 
but these approaches are inefficient because they need to train multiple 
generations synchronously. Therefore, some recent works advocate distilling 
the model within one generation, which take supervision from prior iterations 
\cite{Yang2019,Kim2020}, consistency of distorted data \cite{Xu2019}, 
invariance among intra-class \cite{Yun2020} and the output of deeper portion \cite{Zhang2019}.

These methods only focus on the self-distillation of the classification task. 
In this work, we applied self-distillation to the regression task of depth 
estimation. Different from the method of 
using the whole network to promote sub-networks in \cite{Pilzer2019}, we select 
the optimal disparity map from all output scales as the self-distillation label 
to distill the whole network in one generation.

\noindent
{\bf Data Augmentation.}  
For overfitting, data augmentation is an efficient approach to mitigate 
this drawback by implicitly increasing 
the total amount of training data and teaching models about the 
invariance of the data domain. Common data augmentation methods can 
be summarized into two categories: learnable \cite{Tran2017,Cubuk2019} 
and parameter learning free \cite{Krizhevsky2012,devries2017improved,Zhang2017,Yun2019, Zhong2020}. 
Learnable methods are more universal and work out of the box, while 
the subsequent methods are easier to be implemented and most of them are tailored 
to specific datasets.

Motivated by the fact that the monocular depth estimation model mainly relies on the 
vertical image position and tends to overlook other useful cues, we propose a new parameter 
learning free data augmentation method, called data grafting, to force the model to 
explore more cues.

\section{Method}

We adopt rectified stereo pairs as the input of our self-supervised model 
in training, while only a single image is required to infer depth at 
test time. This kind of self-supervised method is mainly divided into three 
steps. The model $\mathcal{F}:I \rightarrow d \in \mathbb{R}^{H \times W}$, 
that will first estimate the disparity map $d$, which represents the offset of 
the corresponding pixel between the stereo pair, from the target view 
$I \in \mathbb{R}^{C \times H \times W}$. Next, the model will be trained iteratively 
by minimizing the discrepancy between the target view and the view $\tilde{I}$ 
reconstructed from the source view $I'$ with differentiable warping $f_{w}(I',d)$. 
The photometric loss measured with the combination of SSIM \cite{Wang2004} and 
L1 is often adopted to express the discrepancy between the target view and 
the reconstructed view just as:
\begin{equation}
   l_{ph}(d)=l_{ph}(I,\tilde{I}) = \alpha \frac{1-SSIM(I, \tilde{I})}{2}+\beta |I-\tilde{I}|
\label{eq:pe}
\end{equation}
where SSIM() is computed over a $3\times 3$ pixel window, with $\alpha=0.85$ 
and $\beta=0.15$. Finally, the depth map $z \in \mathbb{R}^{H \times W}$ 
will be recovered from $d$, which is outputted by the trained model, with known 
baseline $b$ and focal length $f$ under formula $z=bf/d$. 

In this section, we will introduce the main contributions of this paper in detail. 
The framework pipeline is just shown in Figure \ref{fig:framework}. 

\subsection{Data Grafting} \label{sec:dg}

Lack of data in both quantity and diversity is the first tricky obstacle faced by 
monocular depth estimation, which will damage the generalization ability of the model. 
One of the significant overfitting risks in MDE is the excessive dependence on the vertical 
image position as described in Sec. \ref{sec:intro}. Although data augmentation is the most 
cost-effective and ubiquitous solution, there is almost no relevant research on 
existing self-supervised MDE methods, and only some simple data perturbations such 
as horizontal flipping are used. The reason mainly lies in that self-supervised MDE 
methods generate supervisory signals based on the degree of matching between views, 
which requires strict pixel correspondence (epipolar constraint) to ensure that the 
matching error only comes from the estimated disparity. 
Obviously, the traditional data augmentation method will break this correspondence, 
thereby damaging the performance of the model as shown in our experiments 
in Sec. \ref{sec:abla}. 

However, we note that this restriction is relaxed in the category with stereo pairs 
as input. Because the two views were taken with parallel cameras and rectified, the 
match between them will only occur in the horizontal direction, \eg, panning left 
or right. Therefore, we can do perturbation in the vertical direction to augment 
our data. 

\begin{figure}[t]
   \begin{center}
      \includegraphics[trim={4.3cm 8cm 4.3cm 7cm},clip,width=1.0\linewidth]{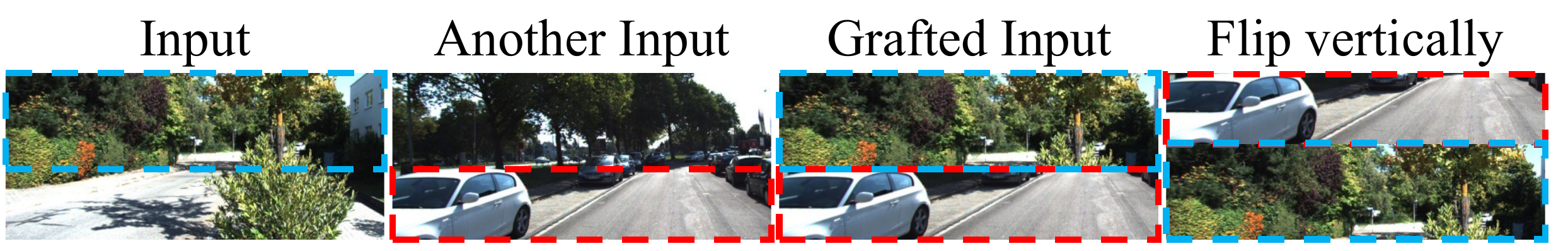}
   \end{center}
   \vspace{-0.3cm}
      \caption{{\bf Illustration of data grafting.} }
   \label{fig:dg}
   \vspace{-0.4cm}
\end{figure}

To this end, we found that grafting two images with different semantics together can 
effectively alleviate the overfitting risk in MDE and encourage the model to better 
utilize the full context of the input without destroying the epipolar constraint. 
We conduct the data grafting within a 
mini-batch, and it is determined by two hyper-parameters: the grafting ratio $r$ and the 
corresponding uniform probability $p$. We reconstruct the input by vertically grafting 
an area with a proportion of $r$ from another input with the probability of $p$, and 
randomly flip these two parts vertically, as shown in Figure \ref{fig:dg}. 
Meanwhile, grafting is not only for the target view, but also for its corresponding 
Depth Hint, which will be introduced in Sec. \ref{sec:tl}, and the source view. 
But each grafting operation can only be performed between the same category, 
\eg, both are target views. And the grafting 
config of all inputs in a batch is the same. The grafting detail of a single 
input is shown in Algorithm \ref{ag:dg}.

\begin{algorithm}
   \label{ag:dg}
   \caption{Data Grafting}
   \LinesNumbered
   \KwIn{Input $I^1$; Another input of the same category randomly sampled from the 
   same batch $I^2$; Shape of input $(c, h, w)$; Random vertical flip factor $flip$.}
   \KwOut{Grafted input $I^1$.}
   Random sampling $r$ from \{0, 0.2, 0.4, 0.6, 0.8\} with the uniform probability of 0.2\;
   \eIf{$r=0$}{
      \KwRet{$I^1$}.
   }{
      $graft\_h=Ceil(h\times r)$\;
      $I^1[:,graft\_h:,:] \leftarrow I^2[:,graft\_h:,:]$\;
      \If{$flip<0.5$}{
         $T = I^1$\;
         $I^1[:,h-graft\_h:,:] \leftarrow T[:,:graft\_h,:]$\;
         $I^1[:,:h-graft\_h,:] \leftarrow T[:,graft\_h:,:]$\;
      }
   }
   \KwRet{$I^1$}.
\end{algorithm}

\subsection{Full-scale Network}

The coarse-to-fine strategy has been proven effective in MDE which continuously refines 
the estimation with iterative warping \cite{Garg2016,Godard2017,Godard2019,Watson2019}. 
The common practice is to output multi-scale 
disparity prediction in the decoder, whose spatial size is incremental. In this 
scenario, the knowledge learned by the encoder is more abstract and general, 
while that in the decoder is more specific to the depth estimation task.

Intuitively, advancing the specialization of depth estimation to the encoder can 
give stronger constraints to the model and further improve its performance. 
Therefore, we extend the traditional multi-scale 
to full-scale, which means that we also add the multi-scale disparity prediction block to 
the encoder. Meanwhile, we insert a residual block, or more precisely an RSU block 
\cite{Qin2020}, between the prediction block and the residual stage in the encoder 
as the bridge to mitigate the impacts between different scales.

\begin{figure}[t]
   \begin{center}
      \includegraphics[trim={6.5cm 3.5cm 8cm 4cm},clip,width=1.0\linewidth]{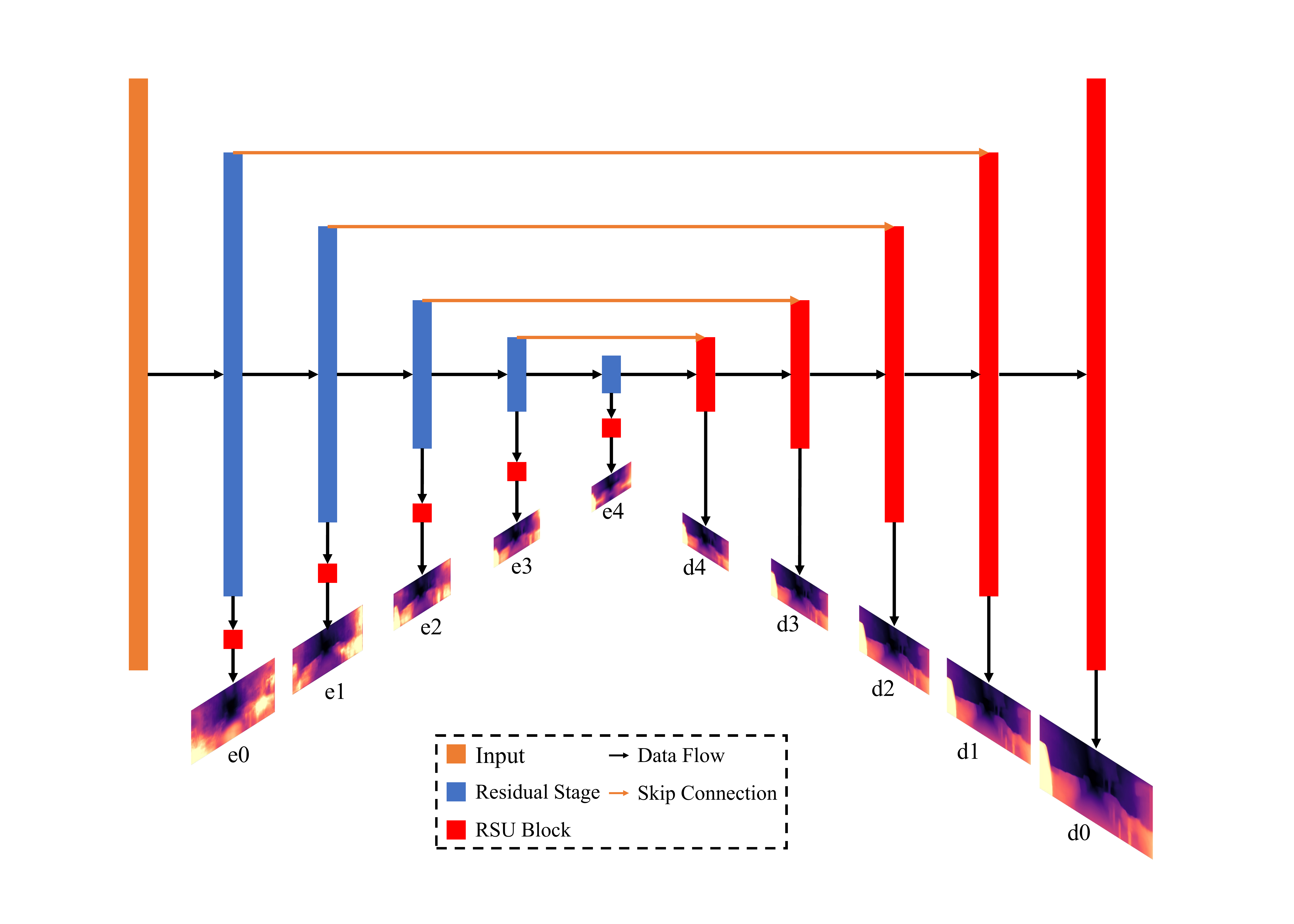}
   \end{center}
      \caption{{\bf Full-scale network.} The ``$e0 \sim e4$'' represents the scale in the 
      encoder and ``$d0 \sim d4$'' represents the scale in the decoder. The spatial 
      size of each scale increases with the decrease of serial number.}
   \label{fig:network}
\end{figure}

Furthermore, just as depicted in Figure \ref{fig:network}, we adopt the RSU block, 
which is more powerful and more lightweight, to construct the decoder to draw the 
representational capacity of our full-scale network. After training, we can 
discard the encoder-scale or even part of the decoder-scale, 
and only retain the largest scale of the decoder, which means that the full-scale 
network will not bring more parameters or computation than the traditional network.

\subsection{Self-distillation} \label{sec:sd}

Self-distillation is an effective way to generate more supervised signals for 
the model, and it is particularly important for self-supervised learning. Here, 
we propose selective post-processing to generate the self-distillation label, 
and with which we create a new loss, termed Self-Distillation Loss $l_{sd}$, 
for the model.

\begin{figure}[t]
   \begin{center}
      \includegraphics[trim={3.2cm 3.2cm 2.5cm 2cm},clip,width=1.0\linewidth]{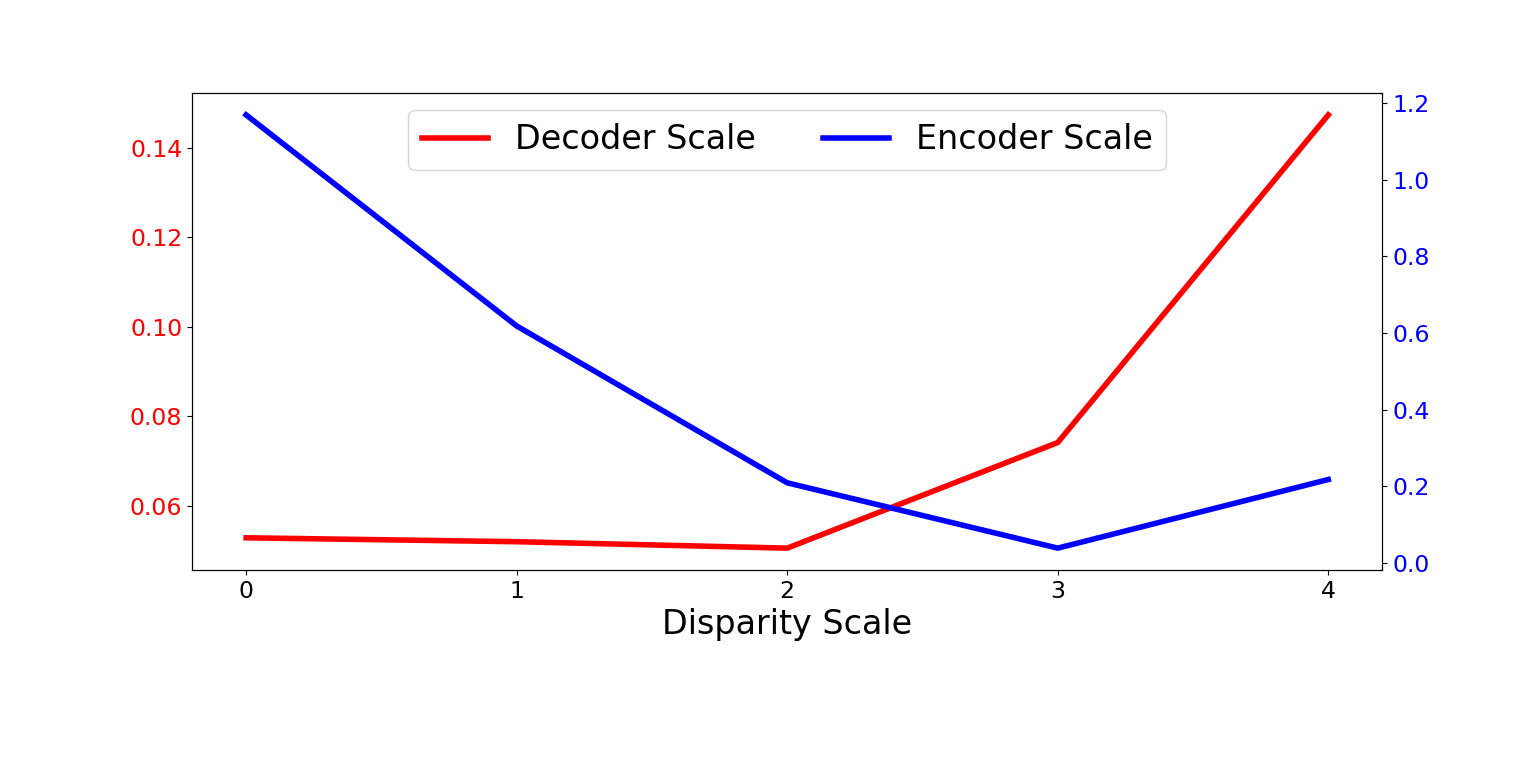}
   \end{center}
   \vspace{-0.2cm}
      \caption{{\bf Precision improvement statistics of SPP result on each scale for all test 
      samples in Eigen split \cite{Eigen2015}.} }
   \label{fig:impro_statistic}
   \vspace{-0.3cm}
\end{figure}

{\bf Selective Post-Processing} aims to filter out the optimal disparity at each pixel 
from multiple disparity scales. Actually, the largest disparity map in the decoder 
that we often output is not always the best at all pixels, as shown in 
Table \ref{tb:scale_compare}. 
Maybe the ``d0'' scale is better at pixel $a$ but the ``d3'' scale is better at pixel $b$. 
Hence, to distinguish the optimal scale at each pixel, we adopt the reconstruction 
error or the photometric loss as our criterion, which is inspired by \cite{Watson2019}. 
Given the full-scale disparity 
maps $\mathrm{D} = [d_{d0}, \dots, d_{d4}, d_{e0}, \dots, d_{e4}]$, 
we will calculate a reconstruction error map for each scale according to 
Equation (\ref{eq:pe}). Then, the self-distillation label of encoder $y_e$ and decoder $y_d$ 
can be constructed based on the assumption that the smaller the error, the better the 
predicted disparity. The detailed procedure of SPP, which 
is the same between the encoder-scale and decoder-scale, is shown in 
Algorithm \ref{ag:spp}. The statistic result in Figure \ref{fig:impro_statistic} shows that 
the SPP can get the most precise results.

\begin{algorithm}
   \label{ag:spp}
   \caption{Selective Post-Processing}
   \LinesNumbered
   \KwIn{The target view $I$; The source view $I'$; Multi-scale disparity maps 
   $\mathrm{D}'$.}
   \KwOut{Self-distillation label $y$.}
   \KwSty{Initialization: $e_{min}=None$}\;
   \For{$d$ \rm{in} $D'$}{
      Upsample $d$ to the same size as $I$\;
      Reconstruct target view $\tilde{I}=f_{w}(I',d)$\;
      Calculate the reconstruction error $e=l_{ph}(I,\tilde{I})$\;
      \eIf{$d=D'[0]$}{
         $y=d$\;
         $e_{min}=e$\;
      }{
         Find all the pixels where $e < e_{min}$\;
         Update $y$ with $d$ at these pixels\;
         Update $e_{min}$ with $e$ at these pixels\;
      }
   }
   \KwRet{$y$}.
\end{algorithm}

{\bf Self-Distillation Loss} is the differenc between the disparity map and the 
self-distillation label for each scale, and it can be modeled as:
\begin{equation}
   l_{sd}(d) = \log{(|y_{c(d)}-d|+1)}
\label{eq:sd_loss}
\end{equation}
where $c(\cdot)$ is used to determine whether $d$ belongs to the 
decoder-scale or the encoder-scale.

\subsection{Training Loss} \label{sec:tl}

Following \cite{Watson2019}, we incorporate the hint loss that has been proven 
effective for thin structures into our model. The Depth Hint $h$ is generated by the 
Semi-Global Matching (SGM) algorithm \cite{Hirschmuller2005,Hirschmuller2008} and be 
consulted only when the reconstruction error can be improved upon. It can be formulated 
for pixel $i$ in each scale as:
\begin{equation}
   l_h(d_{i})= \begin{cases}
      \log{(|h_i-d_{i}|+1)}, & \text {if $l_{ph}(I,\tilde{I}_h)_i<l_{ph}(I,\tilde{I})_i$} \\
      0, &\text{otherwise}
   \end{cases}
\label{eq:final_loss}
\end{equation}
where $\tilde{I}_h$ denotes the reconstructed view with Hint $h$.

Therefore, the final training loss is composed of the average of the three items 
of photometric loss, self-distillation loss and hint loss at each scale:
\begin{equation}
   l=\frac{1}{|\mathrm{D}|} \sum_{d \in \mathrm{D}}(l_{ph}(d) + l_{sd}(d) + l_{h}(d))
\label{eq:final_loss}
\end{equation}


\section{Implementation Details}

We implement our model in PyTorch \cite{paszke2017automatic}. The procedure of calculating 
Depth Hint is the same as that of \cite{Watson2019}. We use Adam \cite{Kingma2015} 
optimizer with the base learning rate of 1e-4 and train the joint loss for 20 epochs. 
Besides our new data 
augmentation approach, we adopted the preprocessing techniques in \cite{Godard2019}. In 
data grafting, we found that the grafting ratio $r=0.2 \times n$, where $n \in\mathbb{N}$ 
and $r<1$, can get the best effect, as shown in Algorithm \ref{ag:dg}. 
Unless otherwise specified, we take ResNet-18 which is pre-trained on ImageNet 
\cite{JiaDeng2009} as the encoder and resize the input to $320 \times 1024$. As for the 
RSU block \cite{Qin2020}, we remove the Batch Normalization layer 
\cite{Ioffe2015} and replace the ReLU \cite{Nair2010} with ELU \cite{Clevert2016} activation. 
More specifically, we take $\mathrm{RSU3} \sim \mathrm{RSU7}$ to construct the decoder's 
layers and the encoder's bridges from minimum scale to maximum scale respectively.

\begin{table*}
   \begin{center}
   \resizebox{0.95\linewidth}{!}{
   \begin{tabular}{|l|c|c|c|c|c|c|c|c|c|c|}
   \hline
   Method & PP & Data & $H\times W$ & \cellcolor[HTML]{E9967A}Abs Rel & \cellcolor[HTML]{E9967A}Sq Rel & \cellcolor[HTML]{E9967A}RMSE & \cellcolor[HTML]{E9967A}RMSE log & \cellcolor[HTML]{B9CEFA}$\delta < 1.25$ & \cellcolor[HTML]{B9CEFA}$\delta < 1.25^2$ & \cellcolor[HTML]{B9CEFA}$\delta < 1.25^3$ \\
   \hline 
   Eigen \etal \cite{Eigen2014} &            & D & $184\times 612$ & 0.203 & 1.548 & 6.307 & 0.282 & 0.702 & 0.890 & 0.890 \\
   Kuznietsov \etal \cite{Kuznietsov2017} &  & DS & $187\times 621$ & 0.113 & 0.741 & 4.621 & 0.189 & 0.862 & 0.960 & 0.986 \\
   Yang \etal \cite{Yang2018Deep}  & \checkmark & $\mathrm{D}^\dagger$S & $256\times 512$ & 0.097 & 0.734 & 4.442 & 0.187 & 0.888 & 0.958 & 0.980 \\
   Luo \etal \cite{Luo2018} &            & $\mathrm{D}^\ast$DS & $192\times 640$ crop & 0.094 & 0.626 & 4.252 & 0.177 & 0.891 & 0.965 & 0.984 \\
   Fu \etal \cite{Fu2018}  &            & D & $385\times 513$ crop & 0.099 & 0.593 & \bf{3.714} & \bf{0.161} & 0.897 & 0.966 & \bf{0.986} \\
   Lee \etal \cite{lee2019big}  &       & D & $352\times 1216$ & \bf{0.091} & \bf{0.555} & 4.033 & 0.174 & \bf{0.904} & \bf{0.967} & 0.984 \\
   \hline
   \hline
   Zhan \etal \cite{Zhan2018} & & MS & $160\times 608$ & 0.135 & 1.132 & 5.585 & 0.229 & 0.820 & 0.933 & 0.971 \\
   Godard \etal \cite{Godard2019} & \checkmark & MS & $320\times 1024$ & 0.104 & 0.775 & 4.562 & 0.191 & 0.878 & 0.959 & 0.981 \\
   Watson \etal \cite{Watson2019} & \checkmark & MS & $320\times 1024$ & \bf{0.098} & 0.702 & 4.398 & 0.183 & 0.887 & 0.963 & \bf{0.983} \\
   Shu \etal \cite{Shu2020} &            & MS & $320\times 1024$ & 0.099 & \bf{0.697} & 4.427 & 0.184 & 0.889 & 0.963 & 0.982 \\
   Lyu \etal \cite{Lyu2021} & & MS & $320\times 1024$ & 0.101 & 0.716 & \bf{4.395} & \bf{0.179} & \bf{0.899} & \bf{0.966} & \bf{0.983} \\
   \hline
   \hline
   Garg \etal \cite{Garg2016} & & S & $188\times 620$ & 0.169 & 1.080 & 5.104 & 0.273 & 0.740 & 0.904 & 0.962 \\
   Godard \etal \cite{Godard2017} & \checkmark & S & $256\times 512$ & 0.138 & 1.186 & 5.650 & 0.234 & 0.813 & 0.930 & 0.969 \\
   Wong \etal \cite{Wong2019} &            & S & $256\times 512$ & 0.133 & 1.126 & 5.515 & 0.231 & 0.826 & 0.934 & 0.969 \\
   Pilzer \etal \cite{Pilzer2019} Teacher &            & S & $256\times 512$ & \bf{0.098} & 0.831 & 4.656 & 0.202 & 0.882 & 0.948 & 0.973 \\
   Chen \etal \cite{Chen2019} & \checkmark & SC & $256\times 512$ & 0.118 & 0.905 & 5.096 & 0.211 & 0.839 & 0.945 & 0.977 \\
   Godard \etal \cite{Godard2019} & \checkmark & S & $192\times 640$ & 0.108 & 0.842 & 4.891 & 0.207 & 0.866 & 0.949 & 0.976 \\
   \underline{Watson \etal} \cite{Watson2019} & \checkmark & S & $192\times 640$ & 0.106 & 0.780 & 4.695 & 0.193 & 0.875 & 0.958 & 0.980 \\
   \bf{Ours} & \checkmark & S & $192\times 640$ & 0.099 & \bf{0.754} & \bf{4.490} & \bf{0.183} & \bf{0.888} & \bf{0.963} & \bf{0.982} \\
   \hline
   Pillai \etal \cite{Pillai2019} & \checkmark & S & $384\times 1024$ & 0.112 & 0.875 & 4.958 & 0.207 & 0.852 & 0.947 & 0.977 \\
   Godard \etal \cite{Godard2019} & \checkmark & S & $320\times 1024$ & 0.105 & 0.822 & 4.692 & 0.199 & 0.876 & 0.954 & 0.977 \\
   \underline{Watson \etal} \cite{Watson2019} & \checkmark & S & $320\times 1024$ & 0.099 & 0.723 & 4.445 & 0.187 & 0.886 & 0.962 & 0.981 \\
   Zhu \etal \cite{Zhu2020} Finetuned & \checkmark & S$\mathrm{C}^\dagger$ & $320\times 1024$ & 0.097 & 0.675 & 4.350 & 0.180 & 0.890 & 0.964 & \bf{0.983} \\
   \bf{Ours} & \checkmark & S & $320\times 1024$ & \bf{0.093} & \bf{0.671} & \bf{4.297} & \bf{0.178} & \bf{0.899} & \bf{0.965} & \bf{0.983} \\
   \hline
   \underline{Watson \etal} \cite{Watson2019} ResNet50 & \checkmark & S & $320\times 1024$ & 0.096 & 0.710 & 4.393 & 0.185 & 0.890 & 0.962 & 0.981 \\
   Zhu \etal \cite{Zhu2020} Finetuned ResNet50 & \checkmark & S$\mathrm{C}^\dagger$ & $320\times 1024$ & \bf{0.091} & \bf{0.646} & 4.244 & 0.177 & 0.898 & \bf{0.966} & \bf{0.983} \\
   {\bf Ours ResNet50} & \checkmark & S & $320\times 1024$ & \bf{0.091} & \bf{0.646} & \bf{4.207} & \bf{0.176} & \bf{0.901} & \bf{0.966} & \bf{0.983} \\
   \hline
   \end{tabular}
   }
   \end{center}
   \caption{{\bf Quantitative results on the KITTI dataset \cite{Geiger2012} using the 
   split of Eigen \etal \cite{Eigen2015}.} Best results in each category are in {\bf bold}. 
   For {\color[HTML]{E9967A}{\bf red}} metrics, lower is better. And higer is better for 
   {\color[HTML]{B9CEFA}{\bf blue}} metrics. Abbreviation in Data column: D refers to methods 
   that are supervised by the ground truth depth, $\mathrm{D}^\dagger$ use auxiliary depth 
   supervision from SLAM, $\mathrm{D}^\ast$ use auxiliary depth supervision from synthetic 
   depth labels, C for supervision from segmentation labels, $\mathrm{C}^\dagger$ for 
   supervision from predicted segmentation labels, S refers to the supervision from stereo 
   images and M for models trained by monocular video. PP represents post-processing 
   \cite{Godard2017}. The underlined model is our baseline. We annotate all the 
   methods that use extra tricks, \eg, fine-tuning and teacher model.}
   \label{tb:compare}
\end{table*}

\section{Experiments}

We first verify the performance of our model on the KITTI dataset \cite{Geiger2012}, 
and perform a comprehensive ablation study on each component. Finally, 
the generalization ability of our model is validated on the NYU-Depth-v2 
dataset \cite{silberman2012indoor}.

\noindent
{\bf KITTI Stereo} was recorded from a driving vehicle and 
contains 42,382 rectified stereo pairs from 61 scenes. To ensure the objectivity of 
comparison, we utilize the Eigen split 
\cite{Eigen2015}, which is composed of 22,600 training image pairs in 32 scenes, and 
697 test pairs in other 29 scenes. We report all seven of the standard metrics 
\cite{Eigen2014} with Garg's crop \cite{Garg2016} and a standard distance cap of 80 
meters \cite{Godard2017}.

\noindent
{\bf NYU-Depth-v2} was captured with a 
Microsoft Kinect sensor and consists of a total 582 indoor scenes. We validate our model 
on the official test set using the same standard metrics as in KITTI.

\begin{figure*}
   \begin{center}
      \includegraphics[trim={0.5cm 6.5cm 0.5cm 6.5cm},clip,width=0.95\linewidth]{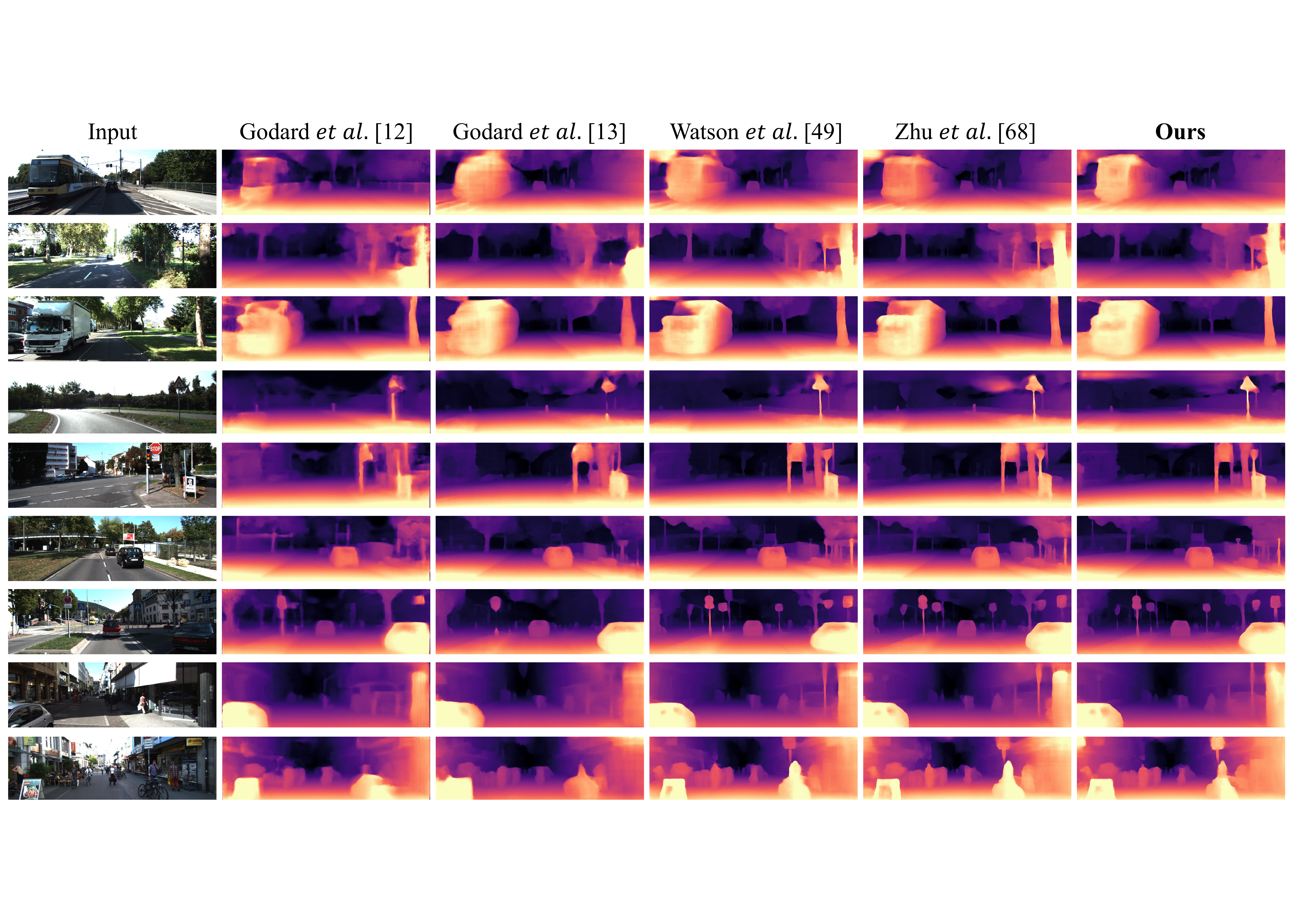}
   \end{center}
   \caption{{\bf Qualitative results.} Our model ({\bf EPCDepth}) in the last column produces the most 
   accurate and sharpest results, especially in challenging areas, \eg, tree trunks, 
   cars, \etc.}
   \label{fig:qualitative}
\end{figure*}

\subsection{Depth Estimation Performance}

We conduct a comprehensive comparison with multifarious methods on the KITTI benchmark to 
verify our depth estimation performance. First of all, we need to emphasize that our 
model is only trained on KITTI stereo data 
and is trick-free. We compare our approach with the recent self-, semi- and fully-supervised 
monocular depth estimation methods in Table \ref{tb:compare}. And the results show that our 
approach outperforms all existing self-supervised methods on all metrics and even some of the 
fully-supervised methods. Our approach of training only on stereo pairs improves 0.013 on 
the $\delta<1.25$ compared to our baseline model \cite{Watson2019}, and this improvement is 
225\% ($=\frac{0.899-0.886}{0.890-0.886}-1$) higher than that of \cite{Zhu2020}, which has finetuned the model and used additional 
constraints. Furthermore, our method is not only outstanding in the category trained with 
stereo images, but also has a major advantage in the category of methods trained with stereo 
video (MS). Even if compared with the best score of each metric in the MS category, our 
approach won out in most metrics. Moreover, we have done more experiments on low-resolution and 
complex backbones to demonstrate the generality and robustness of our model, and the 
consistent performance improvement obtained just proves it.  It's worth noting that we have 
further reduced the gap between full-supervision and self-supervision by nearly 79\% 
($=1-\frac{0.904-0.901}{0.904-0.890}$) compared 
to our baseline \cite{Watson2019}. Besides, the qualitative results in Figure 
\ref{fig:qualitative} show that our model predicts more accurately in challenging areas.

While our model significantly improves the performance of the baseline, it also retains the 
advantages of simple implementation. Each plug-and-play improvement can be easily 
integrated into other models, which is critical for future in-depth studies of monocular 
depth estimation.

\subsection{Ablation Studies} \label{sec:abla}

We perform ablation analysis on the KITTI. The results in Table \ref{tb:ablation} show 
that our full model combining all components has leading performance, and the baseline 
model, without any of our contributions, performs the worst.

\begin{table}
   \begin{center}
   \resizebox{1.0\linewidth}{!}{
   \begin{tabular}{|l||c|c|c|c|}
   \hline
   Data Amount & \cellcolor[HTML]{E9967A}Abs Rel & \cellcolor[HTML]{E9967A}Sq Rel & \cellcolor[HTML]{E9967A}RMSE & \cellcolor[HTML]{B9CEFA}$\delta < 1.25$ \\
   \hline
   w/o DG 100\% & 0.096 & 0.696 & 4.368 & 0.892 \\
   \hline
   Full 20\% & 0.098 & 0.696 & 4.344 & 0.890 \\
   \hline
   Full 50\% & 0.096 & 0.683 & 4.305 & 0.896 \\
   \hline
   Full 100\% & 0.093 & 0.671 & 4.297 & 0.899 \\
   \hline 
   \end{tabular}
   }
   \end{center}
   \caption{{\bf Ablation study on training data amount.} DG refers to data grafting. And 
   the \% means the percentage of data amount.}
   \label{tb:data}
\end{table}

\begin{table}
   \begin{center}
   \resizebox{1.0\linewidth}{!}{
   \begin{tabular}{|l||c|c|c|c|}
   \hline
   Augmentation & \cellcolor[HTML]{E9967A}Abs Rel & \cellcolor[HTML]{E9967A}Sq Rel & \cellcolor[HTML]{E9967A}RMSE & \cellcolor[HTML]{B9CEFA}$\delta < 1.25$ \\
   \hline
   RandErasing \cite{Zhong2020} & 0.115 & 0.992 & 4.987 & 0.858 \\
   \hline
   Cutout \cite{devries2017improved} & 0.106 & 0.830 & 4.753 & 0.874 \\
   \hline
   CutMix \cite{Yun2019} & 0.105 & 0.831 & 4.752 & 0.876 \\
   \hline
   DataGrafting & 0.102 & 0.782 & 4.581 & 0.883 \\
   \hline
   \end{tabular}
   }
   \end{center}
   \caption{{\bf Comparison against other similar augmentation methods.} And the input size is $192\times 640$.}
   \label{tb:augcomp}
\end{table}

\begin{figure}[t]
   \begin{center}
      \includegraphics[trim={2.6cm 1cm 3.7cm 2.3cm},clip,width=1.0\linewidth]{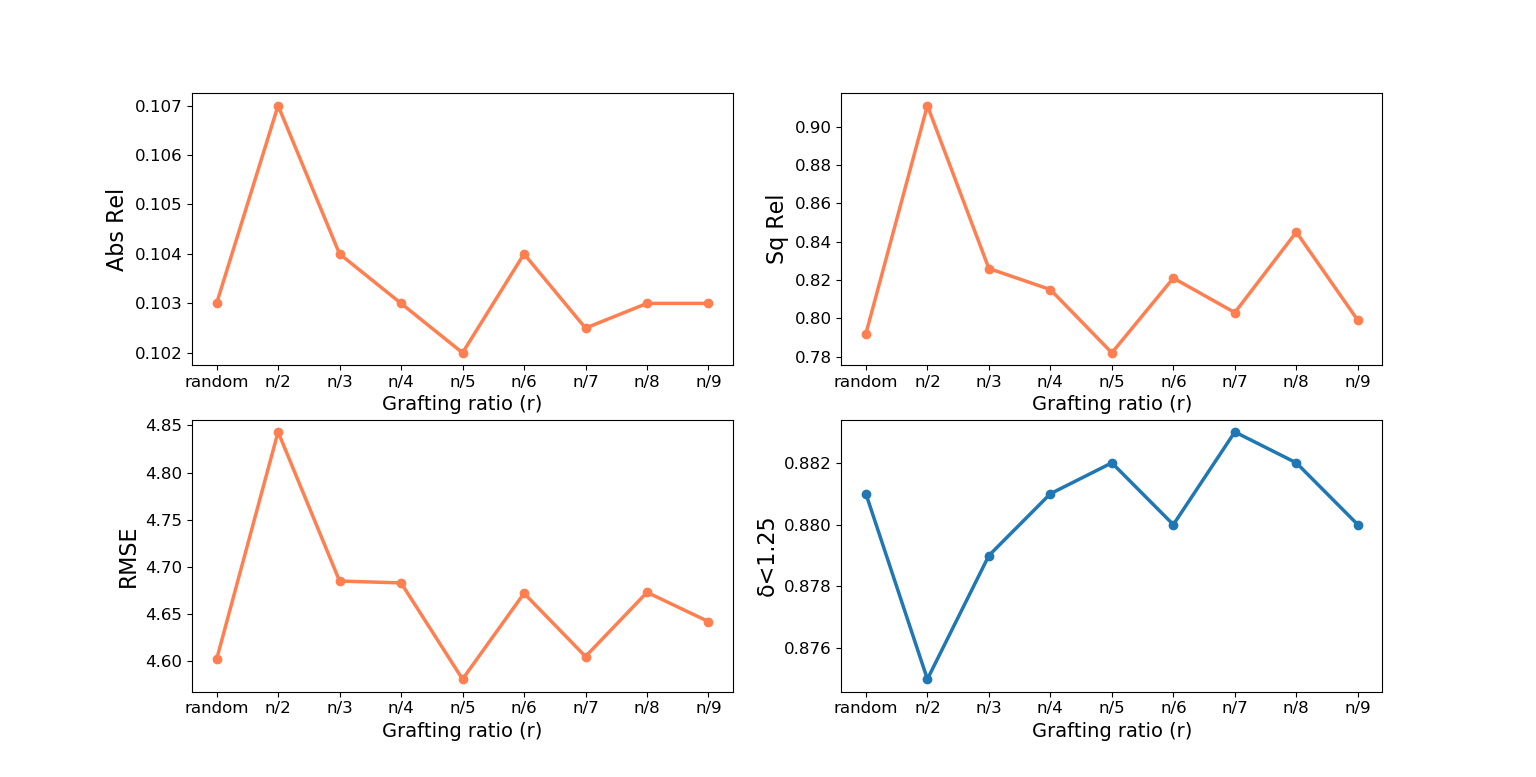}
   \end{center}
      \caption{{\bf Sensitivity analysis of grafting ratio $r$.} The smaller the value, the 
      better in the red line chart, and the worse in the blue.}
   \label{fig:sensitivity}
\end{figure}

\noindent
{\bf Benefits of data grafting.} With data grafting, we can implicitly increase the 
amount of data by $1/p$ times on the basis of our baseline. The 
results in Table \ref{tb:data} show that only 20\% of the data is used to obtain 
competitive performance to the model without data grafting under 100\% of the data, 
which just verifies the strong generalization ability of our model. Moreover, we 
make a comparison with other similar augmentation methods to demonstrate our 
effectiveness in Table \ref{tb:augcomp}. The result just shows that breaking the 
relationship between the depth and the vertical image position with a 
certain probability, which is the uniqueness of data grafting, can allow the model to 
potentially grasp more effective cues. The unsatisfaction of other methods may lie in 
the lack of regularization ability for the vertical image position and the damage of the 
epipolar constraint between views at the edge of the hole. Meanwhile, we 
conduct a sensitivity experiment on the grafting ratio. The resuls in Figure 
\ref{fig:sensitivity} show that the odd setting (\eg n/3) is generally better than 
even setting (\eg n/2), and performs best when $r=n/5$, which indicates that the grafting 
result holding a piece of dominant semantic information is more effective. 

\begin{table}
   \begin{center}
   \resizebox{1.0\linewidth}{!}{
   \begin{tabular}{|l||c|c|c|c|}
   \hline
   Source & \cellcolor[HTML]{E9967A}Abs Rel & \cellcolor[HTML]{E9967A}Sq Rel & \cellcolor[HTML]{E9967A}RMSE & \cellcolor[HTML]{B9CEFA}$\delta < 1.25$ \\
   \hline
   PP & 0.094 & 0.680 & 4.320 & 0.898 \\
   \hline
   SPP & 0.094 & 0.675 & 4.312 & 0.899 \\
   \hline
   SPP separate & 0.093 & 0.671 & 4.297 & 0.899 \\
   \hline
   \end{tabular}
   }
   \end{center}
   \caption{{\bf Ablation study on distillation source.} PP refers to the post-processing 
   result of the largest scale in the decoder.}
   \label{tb:distill_source}
\end{table}

\begin{table}
   \begin{center}
   \resizebox{1.0\linewidth}{!}{
   \begin{tabular}{|l|c|c|c|c|c|c|c|c|c|}
   \hline
   Scale & \cellcolor[HTML]{E9967A}Abs Rel & \cellcolor[HTML]{E9967A}Sq Rel & \cellcolor[HTML]{E9967A}RMSE & \cellcolor[HTML]{B9CEFA}$\delta < 1.25$ & \cellcolor[HTML]{B9CEFA}$\delta < 1.25^2$ \\
   \hline 
   d0 & 0.0925 & 0.671 & 4.297 & {\bf 0.899} & {\bf 0.965} \\
   \hline
   d1 & 0.0922 & 0.668 & 4.292 & {\bf 0.899} & {\bf 0.965} \\
   \hline
   d2 & {\bf 0.092} & {\bf 0.655} & {\bf 4.268} & 0.898 & {\bf 0.965} \\
   \hline
   \end{tabular}
   }
   \end{center}
   \caption{{\bf Quantitative results of different scales.}}
   \label{tb:scale_compare}
\end{table}

\begin{table*}
   \begin{center}
   \resizebox{0.95\linewidth}{!}{
   \begin{tabular}{|l||c|c|c||c||c|c|c|c|c|c|c|}
   \hline
   Method & DG & SD & FS & HR & \cellcolor[HTML]{E9967A}Abs Rel & \cellcolor[HTML]{E9967A}Sq Rel & \cellcolor[HTML]{E9967A}RMSE & \cellcolor[HTML]{E9967A}RMSE log & \cellcolor[HTML]{B9CEFA}$\delta < 1.25$ & \cellcolor[HTML]{B9CEFA}$\delta < 1.25^2$ & \cellcolor[HTML]{B9CEFA}$\delta < 1.25^3$ \\
   \hline
   Baseline & & & & & 0.107 & 0.848 & 4.745 & 0.194 & 0.875 & 0.957 & 0.980 \\
   \noalign{{\color{LightGray}\hrule height 0.4pt}}
   Baseline + DG & \checkmark & & & & 0.102 & 0.782 & 4.581 & 0.188 & 0.883 & 0.960 & 0.981 \\
   \noalign{{\color{LightGray}\hrule height 0.4pt}}
   Baseline + SD & & \checkmark & & & 0.105 & 0.822 & 4.708 & 0.193 & 0.876 & 0.958 & 0.981 \\
   \noalign{{\color{LightGray}\hrule height 0.4pt}}
   Baseline + FS & & & \checkmark & & 0.103 & 0.785 & 4.628 & 0.189 & 0.881 & 0.960 & 0.981 \\
   \hline 
   Baseline HR& & & & \checkmark & 0.101 & 0.758 & 4.497 & 0.187 & 0.886 & 0.962 & 0.982 \\
   \noalign{{\color{LightGray}\hrule height 0.4pt}}
   Baseline HR + DG & \checkmark & & & \checkmark & 0.098 & 0.694 & 4.371 & 0.182 & 0.890 & 0.963 & 0.983 \\
   \noalign{{\color{LightGray}\hrule height 0.4pt}}
   Baseline HR + SD & & \checkmark & & \checkmark & 0.099 & 0.744 & 4.465 & 0.186 & 0.888 & 0.962 & 0.982 \\
   \noalign{{\color{LightGray}\hrule height 0.4pt}}
   Baseline HR + FS & & & \checkmark & \checkmark & 0.097 & 0.701 & 4.364 & 0.182 & 0.892 & 0.963 & 0.982 \\
   \noalign{{\color{LightGray}\hrule height 0.4pt}}
   Full HR w/o FS & \checkmark & \checkmark & & \checkmark & 0.098 & 0.702 & 4.377 & 0.184 & 0.888 & 0.963 & 0.983 \\
   \noalign{{\color{LightGray}\hrule height 0.4pt}}
   Full HR w/o SD & \checkmark & & \checkmark & \checkmark & 0.094 & 0.678 & 4.312 & 0.180 & 0.898 & 0.965 & 0.982 \\
   \noalign{{\color{LightGray}\hrule height 0.4pt}}
   Full HR w/o DG & & \checkmark & \checkmark & \checkmark & 0.096 & 0.696 & 4.368 & 0.182 & 0.892 & 0.963 & 0.982 \\
   \noalign{{\color{LightGray}\hrule height 0.4pt}}
   {\bf Full HR} & \checkmark & \checkmark & \checkmark & \checkmark & \bf{0.093} & \bf{0.671} & \bf{4.297} & \bf{0.178} & \bf{0.899} & \bf{0.965} & \bf{0.983} \\
   \hline
   \end{tabular}
   }
   \end{center}
   \caption{{\bf Ablation resuls amongst variants of our model (EPCDepth) on the KITTI dataset.} DG 
   refers to data grafting, SD refers to self-distillation, FS refers to full-scale and HR 
   refers to high resolution.}
   \label{tb:ablation}
\end{table*}

\noindent
{\bf Benefits of self-distillation.} From Table \ref{tb:scale_compare}, we expect to 
select the optimal scale for each pixel through selective post-processing. The comparison 
results in Table \ref{tb:distill_source} between different label generation methods show 
that the SPP can get more stable improvement and distilling encoder and decoder separately 
is more effective. Meanwhile, we noticed that the magnitude of its performance 
improvement is minimal and is affected by the capacity of the model. But we hope 
that our exploration can open the door to self-distillation in this regression task.

\noindent
{\bf Benefits of full-scale network.} Our full-scale network draws on some advantages of 
the multi-generation strategy, that is to impose more constraints on the model, 
and the results in Table \ref{tb:ablation} 
just prove its power. Furthermore, we explore the effectiveness of the encoder scale, 
RSU blocks and the encoder's bridges respectively, by ablating their effects in 
Table \ref{tb:rsu}. Note that each experiment 
is carried out on the basis of the previous experiment. The continuous performance 
improvement of each module proves their effectiveness. Meanwhile, our full-scale network 
achieves superior performance with {\bf9.88 GFLOPS} at test time, compared to 
{\bf10.1 GFLOPS} of the traditional network \cite{Godard2019,Watson2019,Zhu2020}.

\begin{table}
   \begin{center}
   \resizebox{1.0\linewidth}{!}{
   \begin{tabular}{|l||c|c|c|c|}
   \hline
   Full-Scale & \cellcolor[HTML]{E9967A}Abs Rel & \cellcolor[HTML]{E9967A}Sq Rel & \cellcolor[HTML]{E9967A}RMSE & \cellcolor[HTML]{B9CEFA}$\delta < 1.25$ \\
   \hline
   + Encoder Scale & 0.105 & 0.811 & 4.668 & 0.877 \\
   \hline
   + Bridges & 0.104 & 0.798 & 4.655 & 0.878 \\
   \hline
   + RSU & 0.103 & 0.785 & 4.628 & 0.881 \\
   \hline
   \end{tabular}
   }
   \end{center}
   \caption{{\bf Ablation study on full-scale network}. Conducted by continuously 
   accumulating each module with input size $192\times 640$.}
   \label{tb:rsu}
   \vspace{-0.2cm}
\end{table}

\subsection{Generalizing to NYU-Depth-v2}

\begin{figure}[t]
   \begin{center}
      \resizebox{1.0\linewidth}{!}{
      \includegraphics[trim={1.6cm 24.5cm 1.6cm 15.6cm},clip,width=1.0\linewidth]{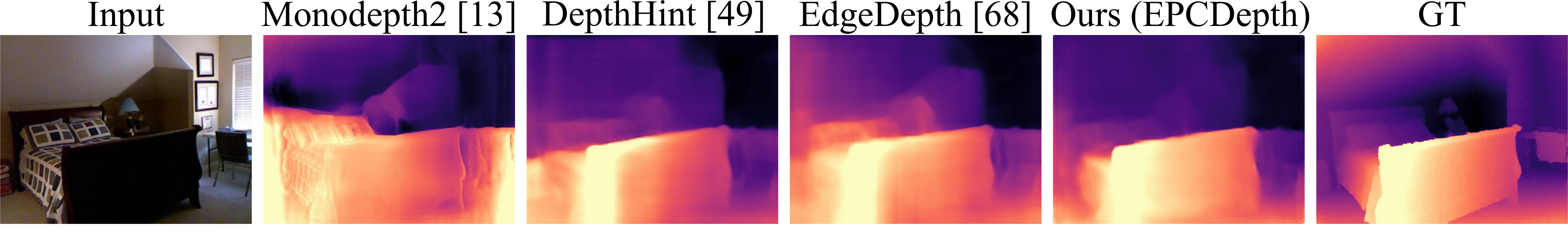}
      }
   \end{center}
   \vspace{-0.2cm}
   \caption{{\bf Qualitative results on the NYU-Depth-v2 dataset.}}
   \label{fig:nyu}
\end{figure}

Since there are no stereo pairs in NYU-Depth-v2 dataset, we train on the KITTI dataset 
and then test on it just as Monodepth \cite{Godard2017} did on Make3D. The preprocessing 
strategy we adopt 
is the same as that of \cite{Yu2020}, and median scaling is applied for all models. The 
results shown in Table \ref{tb:nyu} and Figure \ref{fig:nyu} just verify our strong 
generalization ability. 

\begin{table}[t]
   \begin{center}
   \resizebox{1.0\linewidth}{!}{
   \begin{tabular}{|l|c|c|c|c|c|c|c|c|c|c|}
   \hline
   Method & \cellcolor[HTML]{E9967A}Abs Rel & \cellcolor[HTML]{E9967A}Sq Rel & \cellcolor[HTML]{E9967A}RMSE & \cellcolor[HTML]{E9967A}RMSE log & \cellcolor[HTML]{B9CEFA}$\delta < 1.25$ & \cellcolor[HTML]{B9CEFA}$\delta < 1.25^2$ & \cellcolor[HTML]{B9CEFA}$\delta < 1.25^3$ \\
   \hline
   Monodepth2 \cite{Godard2019} & 0.355 & 0.673 & 1.252 & 0.373 & 0.485 & 0.771 & 0.907 \\
   \hline
   DepthHint \cite{Watson2019} & 0.298 & 0.457 & 1.043 & 0.331 & 0.539 & 0.821 & 0.937 \\
   \hline
   EdgeDepth \cite{Zhu2020} & 0.292 & 0.437 & 1.018 & 0.319 & 0.563 & 0.834 & 0.941 \\
   \hline
   Ours (EPCDepth) & \bf{0.247} & \bf{0.277} & \bf{0.818} & \bf{0.285} & \bf{0.605} & \bf{0.869} & \bf{0.961} \\
   \hline
   \end{tabular}
   }
   \end{center}
   \caption{{\bf Quantitative results on the NYU-Depth-v2 dataset.}}
   \label{tb:nyu}
   \vspace{-0.2cm}
\end{table}

\section{Conclusion}

We extracted the potential capacity of self-supervised monocular depth estimation 
through our novel data augmentation method, 
exploratory self-distillation and efficient full-scale network. The experiments 
demonstrate that our model (EPCDepth) can yield the best performance with the least 
computational cost. In future work, we will try to further improve the performance of 
self-distillation by 
exploring more accurate label generation methods. Besides, applying our contributions 
to other categories, \eg, M, MS and even supervised method, is also a 
potential direction.
\\[5pt]
{\bf Acknowledgements.} Thanks to National Natural Science Foundation of China 
61672063 and 62072013, Shenzhen Research Projects of JCYJ20180503182128089, 
201806080921419290 and RCJC20200714114435057. In addition, we thank the
anonymous reviewers for their valuable comments.

{\small
\bibliographystyle{ieee_fullname}
\bibliography{egbib}
}

\end{document}